\documentclass{article} 
\usepackage{iclr2025_conference,times}

\usepackage{amsmath,amsfonts,bm}

\def\eqref#1{equation~\ref{#1}}

\def\1{\bm{1}}

\DeclareMathAlphabet{\mathsfit}{\encodingdefault}{\sfdefault}{m}{sl}
\SetMathAlphabet{\mathsfit}{bold}{\encodingdefault}{\sfdefault}{bx}{n}

\usepackage{hyperref}
\usepackage{url}
\usepackage[algo2e]{algorithm2e} 
\usepackage{soul}
\usepackage{pifont}
\usepackage{xcolor}
\usepackage{colortbl}
\usepackage[most]{tcolorbox}
\usepackage{algorithm}
\usepackage{pifont}
\usepackage[noend]{algpseudocode}
\usepackage{amsfonts}
\usepackage{mdframed}
\usepackage{xcolor}
\usepackage{bbm}
\usepackage{fdsymbol}
\usepackage{graphicx}
\usepackage{subfig}
\usepackage{mathtools}
\usepackage{amsmath}
\usepackage{amsthm}
\usepackage{listings}
\usepackage{wrapfig}
\definecolor{light_gray}{rgb}{.95,.95,.95}
\definecolor{custompurple}{RGB}{93,0,93}
\definecolor{customorange}{RGB}{255,132,6}
\definecolor{customgold}{RGB}{213,177,52}
\definecolor{customblue2}{RGB}{28,205,188}
\definecolor{no_persona_color}{RGB}{152,226,245}
\definecolor{persona_color}{RGB}{193,167,246}
\usepackage{tcolorbox}
\usepackage{booktabs}
\usepackage{adjustbox}
\usepackage{scalerel}
\usepackage{fontawesome5}
\usepackage{listings}
\usepackage{xcolor}
\makeatletter
\providecommand{\thickapprox}{\mathrel{\thickapprox}}
\makeatother

\lstdefinestyle{llmstyle}{
    backgroundcolor=\color{light_gray},
    basicstyle=\small\ttfamily\color{black},
    keywordstyle=\color{purple},
    commentstyle=\color{gray},
    breaklines=true,
    breakatwhitespace=true,
    frame=single,
    numbers=left,
    numberstyle=\tiny\color{gray},
    rulecolor=\color{black},
    showstringspaces=false,
    tabsize=1,
    breakindent=0pt, 
    escapeinside={(*@}{@*)}  
}

\title{Generative Adversarial Reviews: When LLMs Become the Critic}

\author{
  Nicolas Bougie \\
  Woven by Toyota, Tokyo, Japan \\
  \texttt{nicolas.bougie@woven.toyota} \\
  \And
  Narimasa Watanabe \\
  Woven by Toyota, Tokyo, Japan \\
  \texttt{narimasa.watanabe@woven.toyota}
}

\iclrfinalcopy 
\begin{document}
\maketitle
\begin{abstract}
The peer review process is fundamental to scientific progress, determining which papers meet the quality standards for publication. Yet, the rapid growth of scholarly production and increasing specialization in knowledge areas strain traditional scientific feedback mechanisms. In light of this, we introduce Generative Agent Reviewers (GAR), leveraging LLM-empowered agents to simulate faithful peer reviewers. To enable generative reviewers, we design an architecture that extends a large language model with memory capabilities and equips agents with reviewer personas derived from historical data. Central to this approach is a graph-based representation of manuscripts, condensing content and logically organizing information --- linking ideas with evidence and technical details. GAR’s review process leverages external knowledge to evaluate paper novelty, followed by detailed assessment using the graph representation and multi-round assessment. Finally, a meta-reviewer aggregates individual reviews to predict the acceptance decision. Our experiments demonstrate that GAR performs comparably to human reviewers in providing detailed feedback and predicting paper outcomes. Beyond mere performance comparison, we conduct insightful experiments, such as evaluating the impact of reviewer expertise and examining fairness in reviews. By offering early expert-level feedback, typically restricted to a limited group of researchers, GAR democratizes access to transparent and in-depth evaluation.\end{abstract}

\section{Introduction}

Assessing the quality of research is central to the advancement of scientific discovery. Peer review remains a cornerstone of scientific publication, ensuring that manuscripts meet standards of novelty, rigor, and significance. Although essential, this process faces several challenges, including biases \cite{stelmakh2021prior}, inconsistencies among reviewers \cite{kravitz2010editorial}, and an urgent need for scalable solutions \cite{liu2023reviewergpt}. Estimates suggest that researchers collectively invest millions of hours in reviewing activities annually \cite{AJE2024PeerReview}. Furthermore, access to high-quality feedback remains limited to a small fraction of researchers with established networks. Large language models (LLMs) hold considerable potential in relieving some of these issues in the scientific review process.

Recent breakthroughs in LLMs have shown promise in human behavior modeling by enabling the creation of autonomous agents \cite{hardy2023large,jansen2023employing,argyle2023out,ziems2023can, li2023large}. A growing body of research has explored the application of these LLM-based agents in simulating diverse societal environments  \cite{park2023generative, gao2023s3, törnberg2023simulating, liu2023training, akata2023playing}, with primary emphasis on agents' \emph{cooperation} and \emph{collaboration} behaviors, such as software engineering, playing games, and recommender system evaluation \cite{wu2023autogen, xi2023rise, abdelnabi2023llm, anonymous2024simuser}. However, studies specifically examining the application of LLM-based agents for \textit{academic peer review} remain sparse.

Only a few approaches have explored the use of LLMs as tools to assist researchers at various stages of the scientific workflow, from data analysis to hypothesis generation. Yet, the peer review process remains a particularly challenging domain. For instance, ReviewerGPT has demonstrated how LLMs can identify errors, verify checklists, and select the best version of a paper \cite{liu2023reviewergpt}. Other efforts have shown LLMs capable of reviewing academic manuscripts \cite{liang2024can} and generating creative research ideas \cite{koivisto2023best}. Recently, an attempt has been made to automate the entire scientific endeavor, encompassing research ideation, code writing, experimental execution, results visualization, scientific paper composition, and reviewing \cite{lu2024ai}. Although LLM-generated reviews may be preferred by authors \cite{tyser2024ai}, their ability to predict final paper outcomes still lags behind human experts. In addition, several challenges remain open. These include modeling the intricate relationships between ideas, claims, and technical details in lengthy and complex papers, accurately capturing granular reviewer profiles, and reliably predicting final acceptance outcomes. Addressing these issues is essential to achieving reviewers that match the nuance, diversity, and rigor of human judgment.

We present \textbf{G}enerative \textbf{A}gent \textbf{R}eviewers (GAR), a novel framework that simulates peer reviewers through LLM-based agents. GAR is designed to address two key challenges in the peer review process: (1) providing researchers with early-stage, high-quality feedback across several aspects, such as novelty, significance, technical soundness, and clarity, and (2) predicting acceptance likelihood at major conferences. Each agent is initialized using real-world datasets and equipped with four core modules: profile, memory, novelty, and review modules. The profile module stores traits and historical preferences, including characteristics like \textit{strictness} and \textit{focus areas}, inferred from past reviews via contrastive comparison. The review process begins by constructing a graph-based representation of the manuscript, mapping relationships among ideas, claims, and results. Leveraging this representation alongside retrieved genuine reviews from the memory module, the novelty module assesses the manuscript’s novelty with support from external knowledge. The reviewer module then generates structured feedback and an overall score, which is conducted over multiple rounds, emulating real-world peer review workflows where reviewers provide initial evaluations and refine them in subsequent iterations. Finally, a meta-reviewer agent synthesizes individual reviews to determine the paper’s final decision $\in \{\texttt{[REJECT]},\texttt{[ACCEPT]}\}$.

The contributions of this work are as follows:
\begin{itemize}
    \item We introduce GAR, a framework that automates peer review using LLM agents, providing detailed, on-demand feedback to researchers.
    \item We propose a novel technique to extract reviewer personas from a single review based on contrastive comparison.
    \item We present a novel graph-based representation of academic documents, designed to capture relationships between ideas, claims and results.
    \item We introduce a multi-round review process, wherein the agent generates an initial review and iteratively refines the review based on retrieved information from the memory module and self-assessment.
    \item We propose a meta-reviewer agent that synthesizes individual reviews into a final, consistent decision.
    \item We validate the framework through experiments comparing its performance to human reviewers, highlighting its potential to improve both feedback quality and consistency. We further explore insightful experiments, such as evaluating the impact of reviewer expertise levels and examining fairness in human reviews.
\end{itemize}

\section{Related Work}

The use of artificial intelligence (AI), particularly large language models (LLMs), in scientific discovery has been explored in recent years, with a few studies investigating their potential to replicate and enhance traditional review mechanisms. While generative AI has made strides in areas like molecular modeling \citep{vignac2022digress} and protein structure prediction \citep{abramson2024accurate}, enabling more rapid experimentation, these applications focus primarily on scientific discovery rather than structured assessment and review. Thus, the use of AI in the scientific workflow, especially through large language models (LLMs), remains an emerging area with significant potential.

\subsection{LLMs for Machine Learning Research}
Recent advancements in artificial intelligence have introduced new methodologies that enhance the research process across various domains \cite{xu2021artificial}. For example, \cite{huang2024mlagentbench} introduces a benchmark to evaluate how effectively LLMs can generate code to solve various machine learning tasks. \cite{lu2024discovering} utilize LLMs to propose, build, and test novel algorithms for preference optimization, achieving new performance milestones. Similarly, \cite{liang2024can} demonstrate that LLMs can generate feedback on research papers comparable to human reviewers, while \cite{girotra2023ideas} highlight LLMs’ potential to produce more innovative ideas than humans. Furthermore, \cite{wang2024scimonscientificinspirationmachines, baek2024researchagentiterativeresearchidea} use LLMs to generate research concepts inspired by literature without implementing them, and \cite{wang2024autosurveylargelanguagemodels} automate survey writing through comprehensive literature analysis. Building on these foundations, our method introduces an affordable, on-demand system that can be applied to the field of scientific discovery to assess generated ideas and research papers with human-like depth and rigor.

\subsection{LLMs for Scientific Review and Evaluation}
With the capabilities of LLMs, AI has extended beyond idea generation and experimental design to tasks that closely align with the peer review process \citep{zheng2023large, wang2023scientific, miret2024llms}. By leveraging the natural language understanding and contextual analysis afforded by these models, AI can be utilized to assist in manuscript evaluation by simulating aspects of a human reviewer’s decision-making process \citep{liu2023multi}. Other initial investigations, such as those by Robertson et al. \citep{robertson2023gpt4}, indicated that reviews generated by GPT models align closely with human reviewers’ assessments. By evaluating reviews produced by both humans and LLMs for papers submitted to a prominent machine learning conference, early results illustrated that LLMs could play a valuable role in the peer review process. Further studies \cite{liang2023large} demonstrated that GPT-4’s feedback shared notable similarities with human reviewer comments, with more than half of participants finding the LLM's feedback beneficial, highlighting its increasing impact in this area. Building on these findings, ReviewerMT \cite{tan2024peer} reformulates the peer-review process as a multi-turn, role-based dialogue, encompassing distinct roles for authors, reviewers, and decision makers. 

Recent studies, such as \cite{agentreview2023}, simulate peer review through LLM agents to analyze factors like reviewer biases and decision-making but focus primarily on controlled experiments rather than structured content organization. In contrast, our work incorporates a graph-based representation that clusters key contributions, linking claims, technical details, and evidence to enhance coherence in evaluations. Similarly, \cite{llmreliable2023} examines LLM reliability in review settings, focusing on feedback consistency rather than a comprehensive review simulation. Our approach presents a structured representation to enable detailed multi-dimensional assessments. Research such as \cite{llmasjudge2023} explores LLMs as evaluators in reinforcement learning, assessing their ability to judge AI outputs. However, this study does not address the peer review process comprehensively. Additionally, \cite{llmfeedback2023} shows that LLMs can provide useful feedback on academic documents but lack our structured approach, which systematically links clusters of content to improve interpretability AI-assisted peer review systems like \cite{aiassisted2020} focus on semi-automated tools for improving review speed and accuracy but are limited to assisting human reviewers. In contrast, GAR simulates the full review cycle, from the initial review to the meta-review stages, providing end-to-end automation. 

Recently, MARG \cite{darcy2024marg} has extended this concept by deploying multiple LLMs in a collaborative framework where distinct sections of a paper are allocated to individual agents for focused review, enabling thorough analysis even for extensive documents that surpass the model’s typical context limits. Unlike existing LLM research, which primarily centers on single-pass review generation, our work aims to replicate a holistic review experience by introducing a multi-round approach that connects historical reviews from similar papers with the agent's self-assessment. Finally, research in automated meta-review generation \cite{automatedmeta2023} is limited to synthesizing reviews from existing evaluations. Unlike previous methods, GAR differentiates itself by incorporating a memory-based reviewer equipped with granular personas and a graph-based paper representation to systematically connect evidence with arguments. Through memory-augmented multi-round evaluations and a meta-reviewer to synthesize final decisions, GAR offers a more comprehensive, faithful, and token-efficient simulation of the real-world peer review process.

\section{Methodology}
\begin{wrapfigure}{r}{0.5\textwidth}
    \centering
    \includegraphics[width=0.45\textwidth]{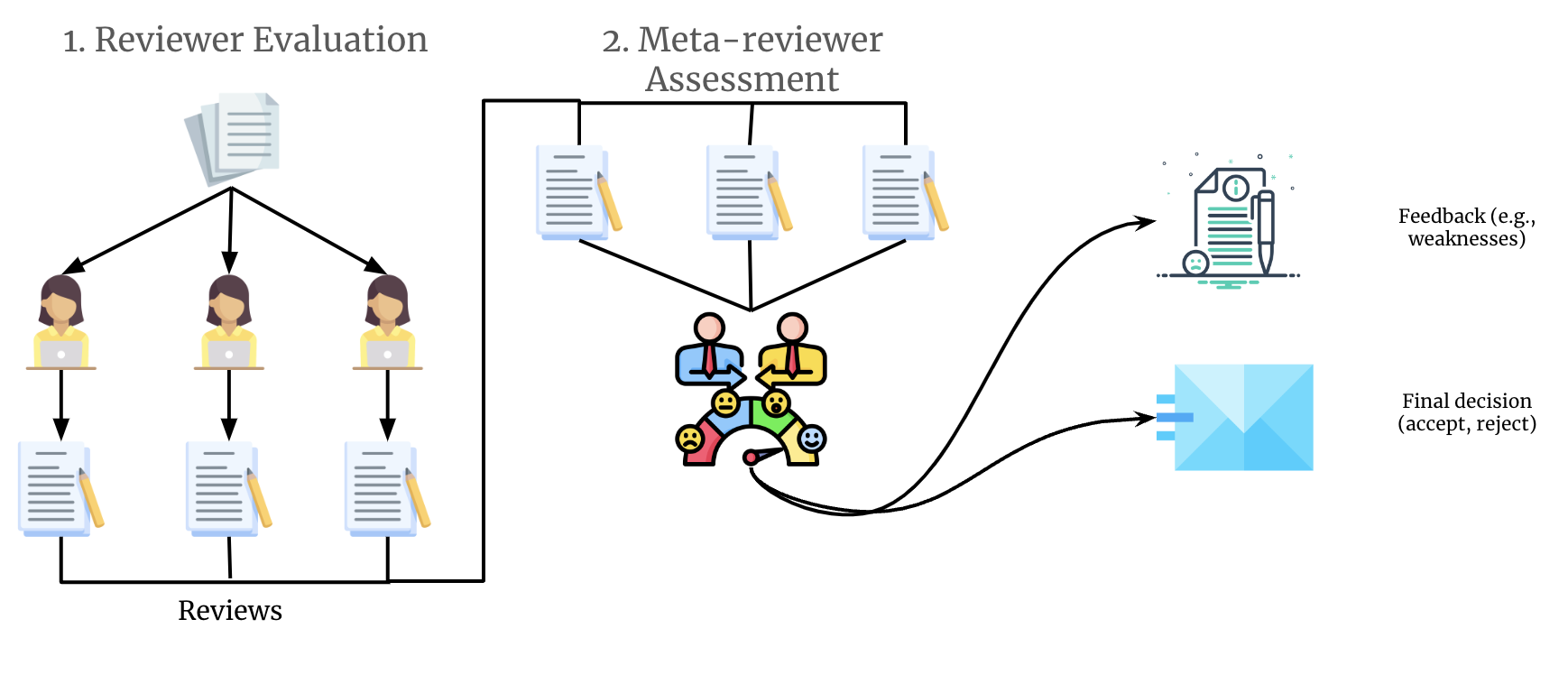}
    \caption{Illustration of GAR framework.}
    \label{fig:simulator}
\end{wrapfigure}
Generative reviewers aim to provide a framework for automated paper review. The system generates numerical scores (soundness, presentation, contribution, overall, confidence), lists weaknesses and strengths, and predicts the acceptance outcome (accept or reject). 

At the center of this approach, we present a graph-based representation of manuscripts, designed to: 1) condense the document length for efficient processing, 2) establish logical links among ideas and novel concepts across different sections, 3) reduce token consumption. Building upon this representation, we introduce the agent's four key modules and describe how a meta-reviewer synthesizes the generated reviews into a final decision.

\textbf{Task Formulation.} Given a paper $p \in \mathcal{P}$ and a reviewer $r \in \mathcal{R}$, let $y_{rp}=1$ denote that reviewer $r$ has reviewer the paper $p$, and subsequently assigned a score $s_{rp}$ with $s_{rp} \in \{1,2,3,4,5,6,7,8,9,10\}$. The average score of each paper $p$ can be represented by $R_{p} = \frac{1}{\sum_{r \in \mathcal{R}} y_{rp}} s_{rp} \cdot y_{rp}$. The simulator’s goal is to faithfully distill the human genuine preferences such as $\hat{y_{rp}}$ and $\hat{s_{rp}}$ of reviewer $r$ for an unseen paper $p$.

\noindent\textbf{Generative Large Language Models.} LLMs are trained to predict the most probable next token $t_{k}$ given the sequence of previous tokens $t_{1} \dots t_{k-1}$ by maximizing the likelihood function $p_{LLM}(t_{k}|t_{1}, \dots , t_{k-1})$. In this work, we use pre-trained LLMs without further finetuning them. Depending on the task, we generate one or more tokens given a task-specific context $c^{(p)}$ that describes the task to the language model and prompts it for an answer. Thus, we obtain generated tokens \textbf{t} by sampling from:
\begin{equation}
    p_{LLM}(t|c^{(p)}) = \prod_{k=1}^{K} p_{LLM}(t_{k}|c_{1}^{(p)}, \dots , c_{n}^{(p)}, t_1, \ldots ,t_{k-1})
\end{equation}
\textbf{Review Process Design.} GAR employs a 4-phase pipeline to simulate the peer review process. \textbf{1. Graph Construction:} In this phase, the manuscript is structured into a knowledge graph that establishes connections between essential ideas, claims, technical details, and results. \textbf{2. Reviewer Selection:} Next, three to six reviewers are selected and their profile module are initialized from historical data. \textbf{3. Reviewer Evaluation:} In this phase, each manuscript undergoes a multi-round evaluation by the independent reviewers. Synthetic reviewers generate comprehensive reviews structured into four sections: significance and novelty, strengths, weaknesses, and suggestions for improvement. This format mirrors the conventional review frameworks utilized by major conferences. Additionally, each reviewer provides an overall score for the paper, assigning a numerical rating on a scale from 1 to 10. \textbf{4. Meta-Review:} Finally, a meta-reviewer compiles the reviews to select the final decision.

\subsection{Graph-Paper Representation}
\label{sec:graph_repr}

As mentioned above, parsing scientific manuscripts is challenging due to their length and the complex relationships among evidence and arguments. Typically, contributions and technical details are introduced in the early sections (e.g., \textit{Introduction} and \textit{Method}), while supporting results are often presented later. This raises several key questions:
\begin{itemize}
    \item \textit{Structuring Information}: How can the diverse elements be effectively organized to enable LLM-based agents to cross-reference and analyze them?
    \item \textit{Processing Lengthy Manuscripts}: Conference papers face efficiency and cost issues with long sequences due to the self-attention mechanism’s limitations\cite{beltagy2020longformer}.
   \item \textit{Reducing Redundancy}: How can redundant claims or findings be minimized to ensure an accurate and thorough assessment?
\end{itemize}
To escape these pitfalls, we introduce a graph-based $\mathcal{G}$ representation that organizes the content of an academic paper $p$ into a structured graph $\mathcal{G}(p)$. Constructing $\mathcal{G}(p)$ consists of the following five steps:
\begin{itemize}
   \item \textbf{Acronym Extraction} The initial step involves identifying and extracting acronyms and their definitions from the manuscript. Acronyms often represent key concepts and terminologies crucial for understanding the paper's content. The LLM parses the title, abstract, and introduction to retrieve a list of acronyms and their corresponding definitions, referred to as $R_{acr}$.    
    \item \textbf{Extraction of Core Elements}: The second step is to identify and extract instances of graph nodes and edges from each chunk of the source paper. Let \(C^{*} = \{c_{1}^{*}, c_{2}^{*}, \dots, c_{n}^{*}\}\) denote the set of chunks in the paper \(p\) (e.g., Introduction, Methods, Results). We leverage a multipart LLM prompt that first identifies all entities in the text, including \textit{ideas}, \textit{claims}, \textit{technical details}, and \textit{supporting evidences}, before identifying all relationships between clearly-related entities, including the source and target entities and a description of their relationship. Each extracted entity \(e \in E\) becomes a node in the graph $\mathcal{G}(p)$, with relationships \(r \in R\) between these elements, such as ``proves'' or ``supports'', represents an edge between nodes. The graph \(\mathcal{G}(p)\) is formally defined as: $\mathcal{G}(p) = (E, R)$.
    \item \textbf{Concept Merging}: To reduce redundancy, we merge nodes that represent the same or similar concepts but are phrased differently across the manuscript. We query the LLM to identify and merge such entities by defining the following merging function: $E' = LLM(\langle Q_{merge}, E, R_{acr} \rangle)$, where $Q_{merge}$ represents the merging prompt, and \( E' \) is the new set of entities after merging the entities $E$. Similarly, if two claims are merged, their technical details and supporting evidence --- edges, will then point to the newly merged entity, $R'$. The updated graph is hence defined as: $\mathcal{G}'(p) = (E', R')$, and $\mathcal{G}(p) \leftarrow \mathcal{G}'(p)$. 
    \item \textbf{Community Detection}: Given the homogeneous undirected weighted graph $\mathcal{G}(p)$, created in the previous step,  a variety of community detection algorithms may be used to partition the graph into communities of nodes with stronger connections to one another than to the other nodes in the graph, with $c$ referring to a community. In all our experiments, we use Leiden \cite{traag2019louvain} to partition the graph into modular communities of closely related nodes \cite{edge2024local}. The modularity \( Q \) is defined as:
    \[
Q = \frac{1}{2m} \sum_{i,j} \left( A_{ij} - \frac{k_i k_j}{2m} \right) \delta(c_i, c_j)
\] where \( A_{ij} \) is the adjacency matrix, \( k_i \) and \( k_j \) are the node degrees, \( m \) is the total number of edges, and \( \delta(c_i, c_j) \) equals 1 if nodes \( i \) and \( j \) are in the same community, 0 otherwise. The Leiden algorithm iteratively optimizes modularity, producing stable and coherent communities of related ideas, claims, and results. This step is essential as it groups nodes into thematically related clusters, allowing the agent to process the manuscript in focused segments. For example, a community might encompass all elements related to a novel loss function, grouping together the key idea, associated claims, technical details, and ablation studies.
\item \textbf{Community-Based Descriptor}: The final step is to create report-like descriptors of each community in the Leiden hierarchy, $\hat{C} = \{\hat{c}_1, \hat{c}_2, \dots, \hat{c}_k\}$. The representation of the nodes and edges in the community serves to query the LLM, which produces a descriptor $\hat{c}_{i}$ representing the community ${c}_{i}$. Each community in the graph is assigned its corresponding descriptor, $\hat{c}_{i} = LLM(\langle Q_{sum}, c_{i}, R_{acr} \rangle)$, where $Q_{sum}$ is a prompt instructing the LLM to describe the community, its structure, and cite the original text as much as possible to mitigate hallucination. These descriptors are attached to the graph $\mathcal{G}(p)$.  
\end{itemize}
The extracted graph $\mathcal{G}(p)$ is passed to the reviewers. For simplicity, we assume that the agents share a unique graph. In future work, we anticipate conditioning the graph extraction with the reviewer's persona.


\subsection{Agent Architecture}
Leveraging this graph-based representation, GAR structures agents in terms of four specialized modules tailored for review scenarios: \textbf{profile}, \textbf{memory}, \textbf{novelty}, and \textbf{review} modules. 

\subsubsection{Profile Module}
The \textbf{profile module} is essential for ensuring the alignment of synthetic agents with the diverse behaviors of genuine reviewers. Drawing inspiration from previous work in user modeling \cite{harper2015movielens,rappaz2021recommendation}, each reviewer persona consists of height core attributes: \textbf{strictness}, \textbf{expertise level}, \textbf{focus areas}, \textbf{evidence focus}, \textbf{open-mindedness}, \textbf{ethic focus}, \textbf{tone}, and \textbf{attention to technical details}.

The following attributes are derived from historical datasets:
\begin{itemize}
    \item \textbf{Strictness} reflects the degree to which a reviewer adheres to high standards in evaluating submissions, ranging from lenient to highly critical.
    \item \textbf{Evidence Focus} describes how much importance the reviewer places on the evidence provided in the submission to support claims, highlighting their emphasis on empirical validation or theoretical soundness.
    \item \textbf{Open-mindedness} measures the reviewer’s willingness to consider unconventional or novel ideas. A higher score indicates more openness to creative methodologies or speculative hypotheses.
    \item \textbf{Tone} refers to the overall style and approach taken by the reviewer in their feedback, ranging from highly critical to constructive.
    \item \textbf{Technical Focus} reflects the extent to which the reviewer is detail-oriented in evaluating the technical correctness and methodological rigor of a submission.
\end{itemize} 
Predicting these characteristics is inherently challenging, as anonymization in blind review limits each reviewer to a single evaluation. To overcome this constraint, we introduce a technique called \textit{contrastive comparison}, which conducts pairwise comparisons across inter-reviewer and intra-reviewer assessments. Specifically, we perform \( N \) comparisons in which the LLM assesses whether the reviewer's review, $\bar{r}$, is stricter than another review from a different paper $\bar{r_{i}}$. To ensure fairness --- acknowledging that stricter reviews are often associated with lower-quality submissions, the LLM is also presented with inter-reviews (anchors) of the same target paper as context. The strictness score \( s^{*}_r \) of the reviewer \( r \) is formally defined as follows:
\begin{equation}
s^{*}_r = \frac{1}{N} \sum_{i=1}^{N} \mathbbm{1}(\text{LLM}\left(r \succ r_{i} \,\big|\, \langle Q_{comp}, \bar{r}, \bar{r}_{int}, \bar{r}_{i} \rangle \right))
\end{equation}
where $Q_{comp}$ is the comparison prompt, $\bar{r}$ is the target reviewer's review, $\bar{r}_{int}$ represents the intra-reviews, and $\bar{r}_{i}$ refers to a review randomly sampled from another reviewer $r_{i}$. Finally, strictness is categorized into $low$, $medium$, and $high$ levels based on percentiles. Similar formulas are derived for the other characteristics. In detail, predicting evidence, open-mindedness, tone, and technical focus involves the same steps with $Q_{comp}$ tailored to each characteristic.

The expertise level and focus areas are assigned as described below:
\begin{itemize}
    \item \textbf{Expertise Level} denotes the depth of the reviewer's knowledge in specific research areas, ranging from novice to expert. The score is derived from real reviews using their confidence scores $\in \{1,2,3,4,5\}$.
    \item \textbf{Focus Areas} defines the primary area of interest for the reviewer, such as theoretical contributions, empirical results, or novel applications. This trait is determined by extracting keywords from past reviews via a prompt $Q_{focus}$. For instance, \texttt{clarity; technical depth; writing quality}, illustrates the focus areas of a reviewer in the ICLR 2023 dataset.
\end{itemize}

\subsubsection{Novelty Module}

The \textbf{novelty module}, which, analogous to a human review, draws upon external knowledge sources to gauge the originality of the manuscript in comparison to prior research. It begins by extracting keywords from the introduction of the targeted paper, which are then employed in a semantic search to retrieve similar papers \cite{ammar-etal-2018-construction}, $\mathcal{B}_{sim}$. Retrieved documents are filtered to ensure that only prior work are included based on the year of submission.

The \textit{title}, \textit{abstract}, and \textit{introduction} of the retrieved papers serve to analyze the extent of innovation, clarity of differences from past contributions, and adequacy of related work citation. The LLM generates a novelty $s_{nov}$ score ranging from 1 (not novel) to 4 (highly novel), accompanied by a concise explanation $e_{nov}$. This scheme can be formalized as: $(s_{nov}, e_{nov}) \leftarrow LLM(\langle Q_{novel}, R_{acr}, \mathcal{B}_{sim}, p \rangle)$, where, with a slight abuse of notation, $p$ denotes the source paper (title, abstract, introduction). To ensure reliable outputs, the LLM is instructed to cite verifiable information, reducing hallucinations or uncertain references. This score $s_{nov} \in \{1,2,3,4\}$ and explanation $e_{nov}$ are subsequently used during paper review to condition reviews.

\subsubsection{Memory Module}
We present a novel \textbf{memory module} designed to support agents. Assuming a benchmark dataset, each academic paper is structured as a graph \( \mathcal{G}(p) = (V, E) \), as detailed in Section \ref{sec:graph_repr}. However, here, we introduce an extra step. Given a community descriptor $\hat{c} \in \hat{C}$, we query the LLM to determine whether the human review mentions $\hat{c}$. If the descriptor is mentioned, the agent is instructed to cite the original review, otherwise, the LLM is prompted to output \texttt{No specific mention was found in the review.}, denoted as $r_{c}$. Following this step, the memory is filled with pairs of community descriptor $\hat{c}$ and their associated reviews as plain text, $\{\hat{c}, r_{c}\}$. All descriptors $\hat{c}$ are embedded using \textit{``mxbai-embed-large''} \cite{li2023angle} and used as index of the memory module, $\mathbf{h}_{\hat{c}}$. 

This memory offers two retrieval schemes, serving at different stages of our framework.
\begin{itemize}
    \item \textbf{Community-level Retrieval:} Retrieve most similar communities and their associated review $\{\hat{c}, r_{c}\}$, straightforwardly using the following similarity function: $\text{sim}(\mathbf{h}_{\textbf{c}}, \mathbf{h}_{\hat{c}'}) = \frac{\mathbf{h}_{\hat{c}}^\top \mathbf{h}_{\hat{c}'}}{\|\mathbf{h}_{\hat{c}}\| \|\mathbf{h}_{\hat{c}'}\|}$, where $\hat{c}$ is the target community descriptor and $\hat{c}'$ represents other communities.
    \item \textbf{Paper-level Retrieval:} Retrieve similar papers based on node and edges overlap. This approach retrieves papers and their associated reviews based on underlying ideas and claims, enabling a more nuanced comparison. Unlike prior studies that rely on direct embedding similarity \cite{specter_cohan_2020}, this scheme compares manuscripts at a descriptor level. The similarly function $\text{sim}_{\text{struct}}$ between two papers $p_{1}$ and $p_{2}$ is expressed:
\begin{align}
        \text{sim}_{\text{struct}}(\mathcal{G}(p_1), \mathcal{G}(p_2)) &= \frac{|\{\hat{c} \in \hat{C}_1 \mid \exists \hat{c}' \in \hat{C}_2 \mathbbm{1}(\text{sim}(\mathbf{h}_{\hat{c}}, \mathbf{h}_{\hat{c}'}) > \tau)\}|}{\max(|\hat{C}_1|, |\hat{C}_2|)}
\end{align}
where \( \tau \) is a scalar that determines whether two communities discuss similar concepts, and $\hat{C}_1$ and $\hat{C}_2$ are community descriptors of $p_{1}$ and $p_{2}$ respectively. 
\end{itemize}

\subsection{Review Module}
Equipping agents with profile and memory modules enables them to exhibit diverse behaviors akin to humans. We enhance the agent’s ability for reasoning via Chain-of-Thought \cite{wei2022chain}. Namely, the review is initiated by the agent processing the paper and generating an \textbf{initial review} $R_{r,0}$ based on its persona and the preliminary novelty assessment $\{s_{nov}, e_{nov}\}$. In this stage, the agent evaluates each \textit{community descriptor} $\hat{C} \in \mathcal{G}(p)$, then outputs numerical scores (soundness, presentation, contribution, overall, confidence), weaknesses and strengths, as well as a preliminary binary decision (accept / reject). The prompt during the initial assessment is formulated as:
\begin{equation}
Q_{r,0} =  \langle Q_{review}, Q_{novelty}, Q_{style}, s_{nov}, e_{nov}, R_{acr}, \hat{C}\rangle
\end{equation}
where the score, accompanied by a summary of supporting arguments, is formatted into plain text and passed on to subsequent review stages. 

To refine the review and enhance its reliability, the agent then engages in \textbf{multi-round refinement}, where the reviewer $r$ at turn $k$ receives the review and thoughts from the previous response $R_{r,k-1}$. Agents are successively presented each community descriptor $\hat{c}$ from the manuscript to review along with the $M$ most similar communities $\hat{c}'_{1}, ..., \hat{c}'_{M}$ retrieved from the memory module and their associated reviews $r_{\hat{c}'_{1}}, ...,  r_{\hat{c}'_{M}}$. To reduce the cognitive workload of reviewers, the agent evaluates communities in blocks of size $\frac{|\hat{C}|}{K}$, where $K$ is the number of review rounds. This retrieval-augmented scheme guides the agent in assessing summaries and discovering potential weaknesses and strengths.

An example prompt block is provided below:
\newtcolorbox{promptblock}{colback=gray!5!white,colframe=gray!75!black,
  fonttitle=\bfseries, title=Prompt Block}
\begin{promptblock}
\textbf{Idea 1:} (\textcolor{orange}{$\hat{c_{1}}$}) Utilizing graph neural networks (GNNs) to model user-item interactions in large-scale recommender systems. The approach claims to enhance scalability and accuracy through advanced message-passing mechanisms. Experiments indicate a 15\% improvement in nDCG@10 compared to baseline collaborative filtering models on the MovieLens dataset.

\bigskip
\textbf{Most Similar Claims:}

\begin{itemize}
    \item \textbf{Similar Claim 1:} (\textcolor{blue}{$\hat{c}'_{1}$}) Implementing dynamic user-item graph construction for scalable recommendations using GNNs... \\
    \textbf{Reviewer comment:} (\textcolor{purple}{$r_{\hat{c}'_{1}}$}) The dynamic graph approach is compelling but could benefit from further comparisons with static graph baselines.\\

    \item \textbf{Similar Claim 2:} (\textcolor{blue}{$\hat{c}'_{2}$}) Applying attention-based GNNs to enhance explainability in recommender systems... \\
    \textbf{Reviewer comment:} (\textcolor{purple}{$r_{\hat{c}'_{2}}$}) The work convincingly demonstrates improved explainability, but additional benchmarks against non-attention models are needed.\\

    \item \textbf{Similar Claim 3:} (\textcolor{blue}{$\hat{c}'_{3}$}) Integrating GNNs with latent factor models to address cold-start issues in recommendation scenarios... \\
    \textbf{Reviewer comment:} (\textcolor{purple}{$r_{\hat{c}'_{3}}$}) The integration with latent factors is innovative, though evaluations on datasets with extreme sparsity could strengthen the claim.\\
\end{itemize}
\textbf{Idea 2:} (\textcolor{orange}{$\hat{c_{2}}$}) ...\\
\textbf{Most Similar Claims:} ...
\end{promptblock}
By drawing inspiration from retrieved exemplars, the agent may decide to add strengths, weaknesses or correct potential mistakes made during the initial review. Thus, the prompt at turn $k$ is formally defined as:
\begin{equation}
    Q_{r,k} =  \langle Q_{check}, R_{r,0}, \overset{K}{\underset{k=1}{\bigcup}}(R_{r,k-1}, (\hat{c}_{1}, ..., \hat{c}_{\frac{|\hat{C}|}{K}}), (\hat{c}'_{1},..., \hat{c}'_{M}, \hat{c}'_{2}, ...), (r_{\hat{c}'_{1}}, ...,  r_{\hat{c}'_{M}}, r_{\hat{c}'_{2}}, ...)\rangle,
\end{equation}
where $R_{r,0}$ denotes the initial review and $R_{r,k-1}$ is the $k$-th review. Given the prompt $Q_{r,k}$, each review agent generates a response $R_{r,k}$, sampled from a probability distribution $R_{r,k} \sim P(\cdot | Q_{r,k})$, as well as the thoughts/rationales behind their choices. The last review is then selected as the final review of reviewer $r$ of the paper $p$.

\subsection{Meta-Reviewer}
After the individual reviews are completed, a meta-reviewer synthesizes the final decision. Each meta-reviewer is equipped with a memory module initialized from genuine meta-reviews, which is leveraged to retrieve the top-$K_{2}$ most similar papers and their meta-reviews. The meta-reviewer is also provided with the scores and reviews provided by the individual reviewers, along with its own preliminarily feedback. It is tasked with synthesizing an overall assessment of the submission, generating a concise but structured summary that highlights both the strengths and weaknesses of the paper, with special attention to the methodological rigor, experimental validation, and impact of the work. This summary aims to consolidate key insights raised by the individual reviewers, presenting a balanced evaluation of the submission.

After $T$ turns of self-reflection, the meta-reviewer generates the final decision $R_{meta}$, following this prompt structure:
\begin{equation}
    Q_{m,t} = \langle Q_{meta}, \hat{r_{1}}, ..., \hat{r_{K_{2}}}  \overset{T}{\underset{t=1}{\bigcup}}(\overline{S_{t}}), \overset{|\mathcal{R}|}{\underset{j=0}{\bigcup}}(R_{j, K})\rangle
\end{equation}
where $\hat{r_{1}}, ..., \hat{r_{K_{2}}}$ represent retrieved meta-reviews based on the similarity between the target paper and manuscript entries in the memory module, $Q_{meta}$ is a meta-review prompt, and $\overline{S_{t}}$ is the meta-reviewer’s summary of dialogues from turn $t$. The final acceptance decision consists of the last review produced at the last round, choosing from the following options: \texttt{ACCEPT (ORAL)}, \texttt{ACCEPT (POSTER)}, or \texttt{REJECT}.

In this work, we compare our method utilizing the above meta-reviewers, $GAR$, with meta-reviewers that accept or reject papers by comparing the review scores with a fixed threshold that reflects real conference acceptance ratio $RAG^{>}$. We discuss the effectiveness these two strategies in the result section (see Sec. \ref{sec:paper_review}).

\section{Environment overview}
Our simulation involves two agent roles: reviewers and meta-reviewers. Each paper is evaluated by a committee comprising three to six reviewers and one meta-reviewer. The paper and review dataset for this simulation is sourced from OpenReview and includes submissions from conferences such as ICLR.

A challenging problem in processing research papers lies in preserving their structural integrity, particularly complex elements like mathematical formulas. Unlike prior work such as AI-Scientist \cite{lu2024ai}, which converts PDFs to plain text and may compromise document formatting, we utilize Nougat \cite{blecher2023nougat} to extract the Markdown (MMD) version of each manuscript, maintaining structural and formatting fidelity.

We argue that experimental results are indispensable for determining a paper’s alignment with publication standards. Hence, we further graft figures into the paper representation. That is, figures that contain empirical findings, such as bar charts, are identified using Molmo-7b \cite{deitke2024molmo}. Next, we prompt GPT-4o to generate detailed captions for these figures, describing the methods being compared, important findings, and key results. Each caption is placed immediately following the original figure title, providing LLM reviewers with direct access to experimental data, enhancing their ability to rigorously evaluate the paper’s technical soundness.

\section{Experiments}
\textbf{Datasets.} We primary conduct the experiments on the ICLR 2023 dataset, which consists of 3,797 papers obtained from Openreview. Each paper was retrieved by at least three reviewers. In some experiments, we also conducted experiments on the ICLR 2022, and NeurIPS 2023 \cite{beygelzimer2021neurips} datasets. 
\\\textbf{Baselines} We compare our method with AI-Scientist \cite{lu2024ai}, OpenReviewer \cite{tyser2024ai}, ReviewerGPT \cite{liu2023reviewergpt}, and AI-Review \cite{chiang2023can}. 
\\\textbf{Implementation.} All agents are powered by the GPT-4o-mini version of ChatGPT \cite{openai2024gpt4technicalreport}. In some experiments, we also use the following state-of-the-art LLMs as the backend of reviewer agents: GPT-4o \cite{openai2024gpt4technicalreport} and Llama-3.1 (8b and 70b) \cite{grattafiori2024llama3herdmodels}. GPT-4o is accessed through a public API, while the Llama-3.1 models, which are open-weight, are deployed via the Ollama framework \citep{ollama}. For each experimental run with Llama-3.1 (8b), we utilize a single NVIDIA A100 40G GPU. Each run on Llama-3.1 (8b) takes approximately 20 minutes, and all results reported are averaged over 20 independent runs to ensure reliability and robustness of the findings. The prompts and other implementation details can be found in the Appendix.

\subsection{LLM vs Human Reviews}
\label{sec:llm_human}

\begin{figure}[tbp]
    \centering
    \includegraphics[width=0.8\textwidth]{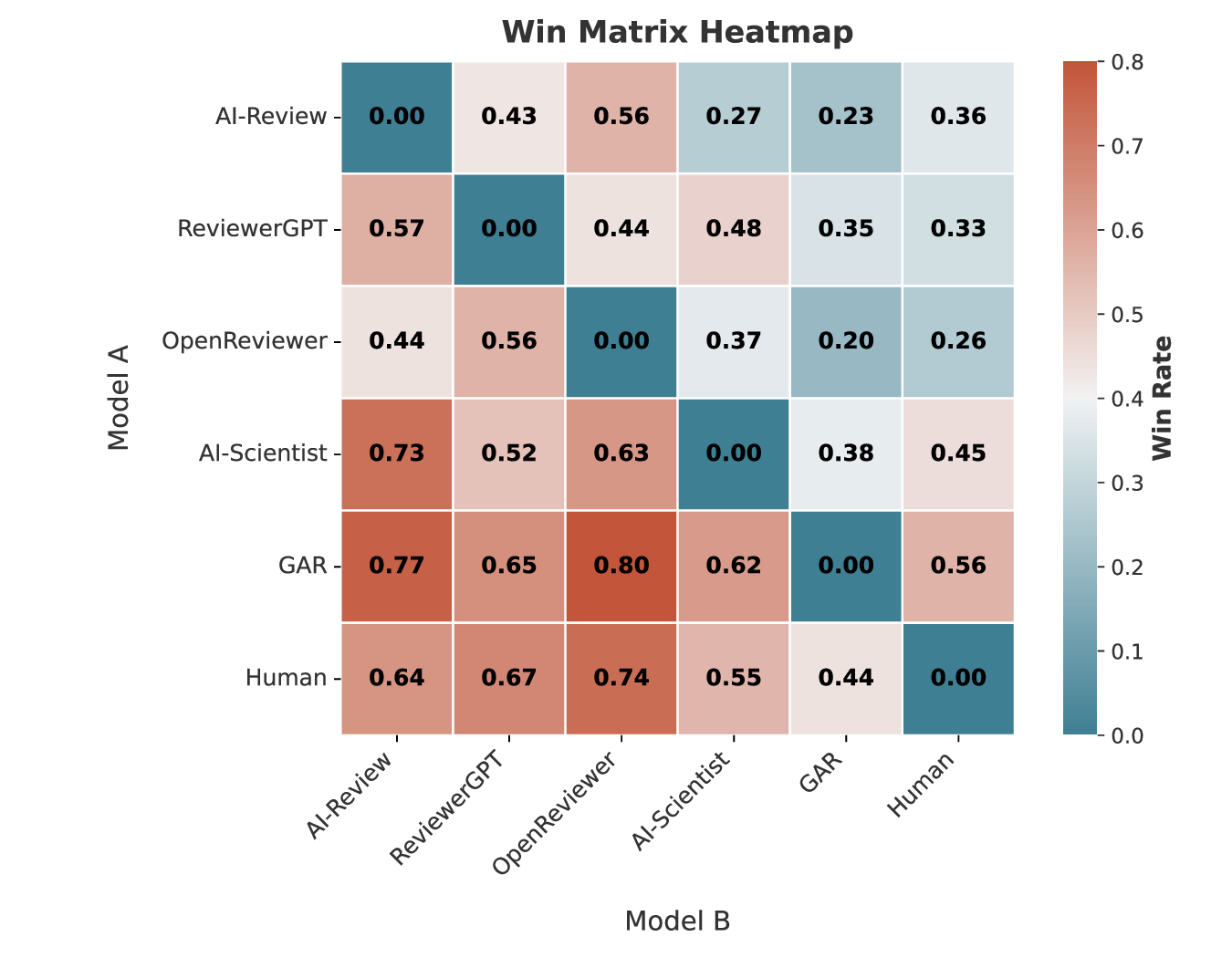}
    \caption{Win rates between six types of reviews (five LLM-generated and a human reviewer) based on GPT-4 preferences.}
    \label{fig:human_reviewer_heatmap}
\end{figure}
In order to assess the quality of reviews generated by large language models, five expert evaluators were given 200 papers, each with two anonymous reviews. As LLM Evaluators \cite{chiang2023can} achieve comparable performance with human evaluators, we use GPT-4o to evaluate the generated reviews. For every paper, two reviewers were randomly assigned among the six possible reviewers: Human, GAR, AI-Scientist, OpenReviewer, ReviewerGPT, and AI-Review. The human reviews were obtained from OpenReview submissions. Evaluators were tasked with selecting their preferred review between the two provided for each paper. 

Therefore, this experiment measures and ranks reviewers based on match outcomes, using a win matrix, coefficients from the Bradley-Terry (BT) model, and logistic regression. The win matrix records the results of matchups between competitors. For \( N \) competitors, the matrix \( W \) is an \( N \times N \) grid where each element \( w_{ij} \) indicates the probability that competitor \( i \) defeats competitor \( j \), calculated as $w_{ij} = \frac{\text{\# wins by } i \text{ over } j}{\text{total matches between } i \text{ and } j}$. The matrix is constructed by processing a list of match results, updating both win counts and total match counts for each competitor pair. The win matrix generated in our experiment is displayed in Figure \ref{fig:human_reviewer_heatmap}.
\begin{table}[tbp]
\centering
\caption{Preference ranking of reviewers based on GPT-4 evaluators. The best results of each model are marked in \textbf{bold} and the second-best results are marked with \underline{underline}.}
\begin{tabular}{ccc}
\toprule
\textbf{Rank} & \textbf{Reviewer} & \textbf{Score} \\
\midrule
\rowcolor{blue!10}
1 & GAR & \textbf{0.684} \\
2 & Human & \underline{0.523} \\
3 & AI-Scientist  & 0.242 \\
4 & ReviewerGPT & 0.000 \\
5 & AI-Review & -0.365 \\
6 & OpenReviewer & -0.632 \\ \hline
\end{tabular}
\label{tab:table_win_rate}
\end{table}

The Bradley-Terry model applies a parametric approach to estimate the relative strengths of competitors through pairwise comparisons. In this model, the probability \( P \) that competitor \( m \) prevails over competitor \( m' \) is given by a logistic function: $P\left(H = \frac{1}{1 + e^{\xi_{m'} - \xi_m}}\right)$, where \( \xi \) represents the vector of BT coefficients, with the constraint \( \xi_1 = 0 \) imposed. These coefficients are derived by minimizing the binary cross-entropy loss over all observed matches, using the following loss function: $\ell(h, p) = -(h \log(p) + (1-h) \log(1-p))$. The optimization task can then be expressed as $\hat{\xi} = \text{argmin}_{\xi} \sum_{t=1}^{T} \ell\left(H_t, \frac{1}{1 + e^{\xi_{A_2} - \xi_{A_1}}}\right)$, while keeping \( \xi_1 = 0 \) to anchor the scale. Once calculated, the BT coefficients \( \xi \) are used to rank competitors, ordering them from strongest to weakest by sorting the \( \xi \) values in descending order, as shown in $\text{Ranked Competitors} = \text{sort}(\xi, \text{descending})$. The results of the Bradley-Terry coefficient estimation, along with the corresponding rankings of reviewers, are displayed in Table \ref{tab:table_win_rate}. 

GAR leads with a score of 0.684, outperforming the human reviewer at 0.523. AI-Scientist and ReviewerGPT achieve scores of 0.242 and 0.000, respectively, while OpenReviewer records the lowest score of -0.632. Upon looking at the responses, GAR generated reviews stem from their depth, rigor, and reduced susceptibility to biases common among individual reviewers, resulting in high preferences. One key factor is the retrieval of relevant reviews for each community descriptor, helping the LLM to identify potential issues or strengths in the proposed method. In contrast, some human reviews are more shallow, often due to reviewers having limited expertise in the field or constrained time for evaluating the manuscript.

\subsection{Human Review Preferences}

\begin{table}[tbp]
\centering
\caption{Human preference ranking of reviewers. The best results of each model are marked in \textbf{bold} and the second-best results are marked with \underline{underline}.}
\begin{tabular}{ccc}
\toprule
\textbf{Rank} & \textbf{Reviewer} & \textbf{Score} \\
\midrule
\rowcolor{blue!10}
1 & GAR & \textbf{0.143} \\
2 & Human & \underline{0.112} \\
3 & AI-Scientist & 0.000 \\
4 & AI-Review & -0.245 \\
5 &  ReviewerGPT & -0.764 \\ 
6 & OpenReviewer & -1.461 \\ \hline
\end{tabular}
\label{tab:table_win_rate_sota}
\end{table}
To evaluate the quality of GAR reviews against prior work, we follow the same experimental settings as in Section \ref{sec:llm_human}, but using human evaluators. As stated above, the evaluators were asked which of the two reviews for each paper they preferred. Ranking of reviewer for our results are shown in Table \ref{tab:table_win_rate_sota}. This experiment evaluates the relative quality of each type of reviewer by having human evaluators make direct, one-on-one comparisons.

Similarly with LLM preferences, GAR achieves a top score of 0.143, higher than the human reviewers. On the other hand, baseline methods underperformed ($\leq0$) compared to human reviewers (0.112). Key factors contributing to GAR’s effectiveness include its memory module and its ability to present similar communities alongside their associated reviewers, which allows for deeper contextualization and relevance in its reviews. These findings underscore GAR’s capability to produce high-quality, contextually informed reviews, positioning it above both conventional AI-based methods and human reviewers. In contrast, most prior LLM-based agents struggle to determine specific criteria for assessment or to identify potential issues effectively, as they do not explicitly retrieve relevant review feedback.

\subsection{Paper Reviewing}
\label{sec:paper_review}
\begin{table}[tbp]
\centering
\caption{Performance comparison of GAR and baselines on three datasets, each consisting of 1,000 papers. The best results for each metric are marked in \textbf{bold}, and the second-best results are marked with \underline{underline}. The improvement of GAR over baselines is statistically significant (measured by student's t-test at $p < 0.05$). (*) Human scores are based on the NeurIPS experiment \cite{beygelzimer2021neurips}.}
\resizebox{1.0\linewidth}{!}{
\begin{tabular}{lcccccc}
\toprule
\textbf{Methods} & \multicolumn{2}{c}{\textbf{NeurIPS}} & \multicolumn{2}{c}{\textbf{ICLR 22}} & \multicolumn{2}{c}{\textbf{ICLR 23}} \\
 & \textbf{Balanced Acc. $\uparrow$} & \textbf{F1 Score $\uparrow$} & \textbf{Balanced Acc. $\uparrow$} & \textbf{F1 Score $\uparrow$} & \textbf{Balanced Acc. $\uparrow$} & \textbf{F1 Score $\uparrow$} \\
\midrule
Human* & \underline{0.66} & \underline{0.49} & \underline{0.66} & \underline{0.49} & \underline{0.66} & \underline{0.49} \\
Random Decision & 0.50 & 0.33 & 0.50 & 0.38 & 0.50 & 0.40 \\
Always Reject & 0.50 & 0.00 & 0.50 & 0.00 & 0.50 & 0.00 \\
\midrule
AI-Scientist & 0.58$\pm$ 0.04 & 0.51$\pm$ 0.06 & 0.65$\pm$ 0.04 & 0.57$\pm$ 0.05 & 0.63$\pm$ 0.05 & 0.55$\pm$ 0.06 \\
OpenReviewer & 0.39$\pm$ 0.05 & 0.39$\pm$ 0.04 & 0.49$\pm$ 0.05 & 0.47$\pm$ 0.05 & 0.50$\pm$ 0.06 & 0.45 $\pm$ 0.05 \\
ReviewerGPT & 0.41$\pm$0.06 & 0.40$\pm$0.05 & 0.54$\pm$0.04 & 0.52$\pm$0.06 & 0.55$\pm$0.07 & 0.51$\pm$0.05 \\
AI-Review & 0.59$\pm$0.04 & 0.49$\pm$0.05 & 0.64$\pm$0.06 & 0.55$\pm$0.04 & 0.61$\pm$0.06 & 0.53$\pm$0.07 \\
\midrule
\rowcolor{blue!10}
GAR & \underline{0.64$\pm$0.05} & \underline{0.61$\pm$0.04} & \underline{0.68$\pm$0.03} & \underline{0.66$\pm$0.05} & \underline{0.66$\pm$0.04} & \underline{0.60$\pm$0.04} \\
\rowcolor{blue!10}
GAR$^{>}$ & \textbf{0.68$\pm$0.05} & \textbf{0.62$\pm$0.05} & \textbf{0.71$\pm$0.04} & \textbf{0.67$\pm$0.06} & \textbf{0.70$\pm$0.05} & \textbf{0.69$\pm$0.05} \\
\bottomrule
\end{tabular}
}
\label{tab:gar_results}
\end{table}

To evaluate the effectiveness of our LLM-powered review system, we compared its decisions against a ground truth dataset comprised of 1,000 papers from the NeurIPS 23, ICLR 22, and ICLR 23 submissions. The remaining reviews (e.g., 2,797 for ICLR 23) in each dataset were utilized to initialize the memory module. Table \ref{tab:gar_results} summarizes the performance metrics. Our method outperforms previous state-of-the-art methods, including AI-Scientist (0.54) with an average f1 score of 0.66. This f1 score is significantly higher than the 0.49 achieved by human reviewers in the NeurIPS 2023 consistency study \cite{beygelzimer2021neurips}, as indicated by paired t-tests at 95\% confidence level (p $<$ 0.002). In GAR$^{>}$, we set the decision threshold at a score of 6, aligned with the "Weak Accept" category from ICLR's review standards, and compare this threshold-based reviewers with vanilla GAR. As a result, GAR$^{>}$ exhibits significantly superior performance compared to GAR. We attribute GAR's improvements to the use of a profile module augmented with granular information (e.g., strictness, focus area) obtained via contrastive matching. On the other hand, prior LLM-based methods struggle to assess complex and lengthy paper, resulting in lower than humans performance. The present study alleviates this pitfall by leveraging the graph representation of manuscripts and the memory module to retrieve relevant genuine reviews.

\subsection{Experiment: Reviewer Persona Selection}

\begin{table}[tbp]
\centering
\caption{Ablation study of GAR on three datasets, each consisting of 1,000 papers. Lines 3-5 report results for different approaches to select reviewers' persona. Line 6 highlights the performance without memory module.}
\resizebox{1.0\linewidth}{!}{
\begin{tabular}{lcccccc}
\toprule
\textbf{Methods} & \multicolumn{2}{c}{\textbf{NeurIPS}} & \multicolumn{2}{c}{\textbf{ICLR 22}} & \multicolumn{2}{c}{\textbf{ICLR 23}} \\
 & \textbf{Balanced Acc. $\uparrow$} & \textbf{F1 Score $\uparrow$} & \textbf{Balanced Acc. $\uparrow$} & \textbf{F1 Score $\uparrow$} & \textbf{Balanced Acc. $\uparrow$} & \textbf{F1 Score $\uparrow$} \\
\midrule
GAR (random persona) & 0.59$\pm$0.05 & 0.57$\pm$0.05 & 0.60$\pm$0.07 & 0.68$\pm$0.06 & 0.61$\pm$0.03 & 0.61$\pm$0.06  \\
GAR$^{>}$ (random persona) & 0.62$\pm$0.05 & 0.59$\pm$0.04 & 0.63$\pm$0.04 & 0.69$\pm$0.05 & 0.64$\pm$0.06 & 0.65$\pm$0.06  \\
GAR (historical persona) & 0.65$\pm$0.05 & 0.60$\pm$0.04 & 0.65$\pm$0.06 & 0.64$\pm$0.04 & 0.69$\pm$0.06 & 0.66$\pm$0.05 \\
GAR$^{>}$ (historical persona) & \underline{0.68$\pm$0.05} & \underline{0.62$\pm$0.05} & 0.71$\pm$0.04 & \underline{0.67$\pm$0.06} & 0.70$\pm$0.05 & \underline{0.69$\pm$0.05} \\
GAR (NN selected) & 0.66$\pm$0.07 & 0.61$\pm$0.04 & \underline{0.72$\pm$0.05} & 0.66$\pm$0.05 & \underline{0.72$\pm$0.06} & 0.67$\pm$0.05  \\
GAR$^{>}$ (NN selected) & \textbf{0.70$\pm$0.06} & \textbf{0.63$\pm$0.05} & \textbf{0.74$\pm$0.05} & \textbf{0.69$\pm$0.04} & \textbf{0.74$\pm$0.05} & \textbf{0.70$\pm$0.06}  \\
\midrule
GAR (w/o memory) & 0.54$\pm$0.04 & 0.54$\pm$0.06 & 0.61$\pm$0.05 & 0.53$\pm$0.04 & 0.61$\pm$0.05 & 0.52$\pm$0.06  \\
GAR$^{>}$ (w/o memory) & 0.59$\pm$0.04 & 0.54$\pm$0.05 & 0.65$\pm$0.04 & 0.58$\pm$0.06 & 0.64$\pm$0.05 & 0.54$\pm$0.05  \\
\midrule
\rowcolor{blue!10}
GAR & 0.64$\pm$0.05 & 0.61$\pm$0.04 & 0.68$\pm$0.03 & 0.66$\pm$0.05 & 0.66$\pm$0.04 & 0.60$\pm$0.04 \\
\rowcolor{blue!10}
GAR$^{>}$ & \underline{0.68$\pm$0.05} & \underline{0.62$\pm$0.05} & 0.71$\pm$0.04 & \underline{0.67$\pm$0.06} & 0.70$\pm$0.05 & \underline{0.69$\pm$0.05} \\
\bottomrule
\end{tabular}
}
\label{tab:review_strategy}
\end{table}

Reviewer personas shape the tone and content of paper evaluations. We compare three approaches for selecting reviewer personas: 
\begin{itemize}
    \item \textbf{Random Persona Selection}: Reviewer profiles are randomly selected among possible personas. This approach serves as the baseline, representing the scenario where the persona is not known in advance.
    \item \textbf{Historical Data Initialization}: Reviewer personas are initialized using historical data from prior reviews, matching the profile of a reviewer based on their past decisions and subject matter expertise. This setting replicates the reviews collected in real-world datasets.
    \item \textbf{Neural Network-Optimized Persona}: In this approach, a small neural network is trained to select reviewer personas that maximize the paper acceptance rate.
\end{itemize}

Results (Table \ref{tab:review_strategy}) depict that \textit{Neural Network-Optimized Persona} and \textit{historical data initialization} produce the most consistent results, achieving a 0.74 and 0.70 on ICLR 23, respectively. This suggests that further tuning of the network could reduce over-acceptance of borderline papers, while maintaining a strong acceptance rate for high-quality submissions. On the other hand, the \textit{random selection} method, while less accurate, performs at 0.64. For the \textit{random selection} setting, inconsistencies arise due to misaligned reviewer traits, such as focus areas or strictness levels, which can lead to decisions that diverge significantly from those of real reviewers. These findings underscore the importance of the profile module, and its impact on achieving human-like evaluations.

\subsection{Effect of Paper Representation}
\definecolor{myGreen}{RGB}{11,84,0}
\definecolor{myblue}{RGB}{2,125,181}
\definecolor{myRed}{RGB}{255,123,122}

\begin{table}[tbp]
\centering
\caption{Effect of paper representation on the performance for predicting the acceptance of manuscripts. Asterisks (*) denote statistically significant improvements over the best baseline (t-test at $p<0.05$).}
\label{tab:performance_metrics}
\resizebox{1.0\linewidth}{!}{
\begin{tabular}{lcccccc}
\toprule
\textbf{Methods} & \multicolumn{2}{c}{\textbf{NeurIPS}} & \multicolumn{2}{c}{\textbf{ICLR 22}} & \multicolumn{2}{c}{\textbf{ICLR 23}} \\
 & \textbf{Balanced Acc. $\uparrow$} & \textbf{F1 Score $\uparrow$} & \textbf{Balanced Acc. $\uparrow$} & \textbf{F1 Score $\uparrow$} & \textbf{Balanced Acc. $\uparrow$} & \textbf{F1 Score $\uparrow$} \\
\midrule
GAR (\textcolor{myRed}{$\vardiamondsuit$}) & \underline{0.64$\pm$0.05} & \underline{0.61$\pm$0.04} & \underline{0.68$\pm$0.03} & \underline{0.66$\pm$0.05} & \underline{0.66$\pm$0.04} & \underline{0.60$\pm$0.04} \\
\rowcolor{blue!10}
GAR$^{>}$ (\textcolor{myRed}{$\vardiamondsuit$}) & \textbf{0.68$\pm$0.05} & \textbf{0.62$\pm$0.05} & \textbf{0.71$\pm$0.04} & \textbf{0.67$\pm$0.06} & \textbf{0.70$\pm$0.05} & \textbf{0.69$\pm$0.05} \\
\midrule
GAR (\textcolor{myblue}{$\spadesuit$}) & 0.60$\pm$0.06 & 0.58$\pm$0.05 & 0.65$\pm$0.05 & 0.63$\pm$0.04 & 0.64$\pm$0.05 & 0.57$\pm$0.06 \\
\rowcolor{blue!10}
GAR$^{>}$ (\textcolor{myblue}{$\spadesuit$}) & 0.63$\pm$0.05 & 0.60$\pm$0.06 & 0.67$\pm$0.04 & 0.63$\pm$0.05 & 0.63$\pm$0.05 & 0.57$\pm$0.05  \\
\midrule
GAR (\textcolor{persona_color}{$\varheartsuit$}) & 0.55$\pm$0.05 & 0.52$\pm$0.04 & 0.63$\pm$0.05 & 0.56$\pm$0.06 & 0.60$\pm$0.05 & 0.53$\pm$0.05  \\
\rowcolor{blue!10}
GAR$^{>}$ (\textcolor{persona_color}{$\varheartsuit$}) & 0.58$\pm$0.05 & 0.53$\pm$0.04 & 0.65$\pm$0.06 & 0.59$\pm$0.04 & 0.61$\pm$0.06 & 0.64$\pm$0.06  \\
\bottomrule
\end{tabular}
}
\label{tab:f1_scores}
\end{table}
To validate the significance of our graph-based representation, we compare RAG's effectiveness in predicting paper acceptance using three input types: (1) community descriptors \textcolor{myRed}{$\vardiamondsuit$}, (2) node-edge representations \textcolor{myblue}{$\spadesuit$}, and (3) raw text \textcolor{persona_color}{$\varheartsuit$}. The experiment results, illustrated in Table \ref{tab:f1_scores}, demonstrate that different manuscript representation methods significantly influence the F1-scores achieved in the review process. Our analysis shows that structured representations, particularly graphs and community-based descriptors, yield higher F1-scores compared to raw text. These structured formats enhance the extraction and alignment of key concepts, leading to more precise and contextually aware assessments. In contrast, raw text lacks such structured organization, resulting in lower F1-scores due to its reliance on linear narrative, which may increase the cognitive load for reviewers. This observation aligns with existing research in natural language processing, which underscores the value of structured representations in enhancing information extraction and interpretation \cite{liu2021learning, schneider2022decade}.

\subsection{Review Scores Across Different Models}
\begin{figure}[tbp]
    \centering
    \includegraphics[width=1.0\textwidth]{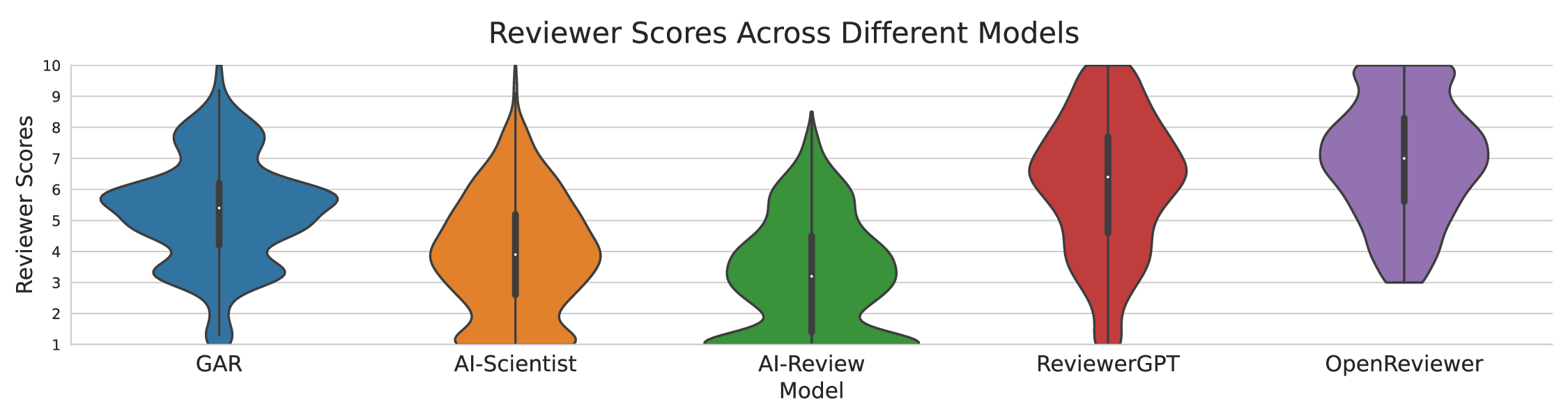}
    \caption{Violin plots showing the distribution of scores generated by several AI-generated models for ICLR 23 papers. Scores on the y-axis refer to paper ratings, which range from 1 (Strong Reject) to 10 (Weak Accept).}
    \label{fig:review_score_distribution}
\end{figure}

This experiment validates the effectiveness of five LLM-powered reviewers in scoring ICLR 23 papers: GAR, AI-Scientist, AI-Review, ReviewerGPT, and OpenReviewer. Figure \ref{fig:review_score_distribution} shows violin plots of score distributions aligned with ICLR ratings, allowing for an assessment of each model’s alignment with human review standards. AI-Scientist exhibits a concentrated distribution around 4, indicating moderate alignment, while GAR showed greater variability. ReviewerGPT and OpenReviewer skewed higher, suggesting a higher inclination to accept manuscripts. On the other hand, GAR demonstrates consistent scores between 1 and 10, with the ability to accept high-quality papers (score $>$7) while strongly rejecting papers ($\leq$3) that do not meet the quality standards for publication.

\subsection{Effect of Reviewer Expertise Level}
The effect of reviewer expertise on the acceptance likelihood and feedback quality is unclear. Thus, we investigate how varying levels of simulated reviewer expertise impact the quality and consistency of reviews. We set the persona of reviewers with the following expertise levels: a \textit{Novice Reviewer} with limited knowledge, an \textit{Intermediate Reviewer} with general familiarity with the field, and an \textit{Expert Reviewer} with deep expertise in the research domain. As a baseline, we also report the results of GAR that assigns the expertise level based on genuine values from OpenReview.

The results, summarized in Table \ref{tab:expertise}, reveal that reviewers with higher expertise levels consistently outperformed those with lower expertise. That is, expert reviewers achieve the highest accuracy and consistency, approaching the level expected of human experts. Notably, novice and intermediate reviewers also produced reasonably accurate assessments, but their performance lagged behind that of expert-level reviewers. This suggests that emulating expertise levels in simulated reviewers improves the quality of automated reviews, supporting the use of calibrated expertise to enhance the reliability and value of LLM-based reviews. Analysis of reviews revealed that reviewers with higher expertise tend to produce more detailed assessments, while novice reviewers focus on more shallow considerations, including asking authors to add more experiments and implication of the research.

\begin{table}[tbp]
\centering
\caption{Performance for predicting the acceptance of manuscripts with varying levels of review expertise.}
\label{tab:expertise}
\resizebox{1.0\linewidth}{!}{
\begin{tabular}{lcccccc}
\toprule
\textbf{Methods} & \multicolumn{2}{c}{\textbf{NeurIPS}} & \multicolumn{2}{c}{\textbf{ICLR 22}} & \multicolumn{2}{c}{\textbf{ICLR 23}} \\
 & \textbf{Balanced Acc. $\uparrow$} & \textbf{F1 Score $\uparrow$} & \textbf{Balanced Acc. $\uparrow$} & \textbf{F1 Score $\uparrow$} & \textbf{Balanced Acc. $\uparrow$} & \textbf{F1 Score $\uparrow$} \\
\midrule
Novice Reviewers & 0.57$\pm$0.05 & 0.55$\pm$0.05 & 0.61$\pm$0.04 & 0.59$\pm$0.05 & 0.59$\pm$0.04 & 0.57$\pm$0.05 \\ 
Novice Reviewers$^>$ & 0.59$\pm$0.05 & 0.56$\pm$0.05 & 0.63$\pm$0.04 & 0.60$\pm$0.05 & 0.61$\pm$0.04 & 0.58$\pm$0.05 \\ 
\midrule
Intermediate Reviewers & 0.60$\pm$0.05 & 0.58$\pm$0.05 & 0.64$\pm$0.04 & 0.62$\pm$0.05 & 0.62$\pm$0.04 & 0.60$\pm$0.05 \\
Intermediate Reviewers$^>$ & 0.62$\pm$0.05 & 0.59$\pm$0.05 & 0.66$\pm$0.04 & 0.64$\pm$0.05 & 0.64$\pm$0.04 & 0.62$\pm$0.05 \\
\midrule
Expert Reviewers & 0.62$\pm$0.07 & 0.60$\pm$0.04 & 0.65$\pm$0.05 & 0.62$\pm$0.04 & 0.64$\pm$0.05 & 0.61$\pm$0.05 \\
Expert Reviewers$^>$ & 0.66$\pm$0.05 & 0.61$\pm$0.03 & 0.68$\pm$0.05 & 0.63$\pm$0.05 & 0.67$\pm$0.05 & 0.62$\pm$0.04 \\
\midrule
\rowcolor{blue!10}
GAR & \underline{0.64$\pm$0.05} & \underline{0.61$\pm$0.04} & \underline{0.68$\pm$0.03} & \underline{0.66$\pm$0.05} & \underline{0.66$\pm$0.04} & \underline{0.60$\pm$0.04} \\
\rowcolor{blue!10}
GAR$^{>}$ & \textbf{0.68$\pm$0.05} & \textbf{0.62$\pm$0.05} & \textbf{0.71$\pm$0.04} & \textbf{0.67$\pm$0.06} & \textbf{0.70$\pm$0.05} & \textbf{0.69$\pm$0.05} \\
\bottomrule
\end{tabular}
}
\end{table}

\subsection{Review Scores Across Different LLMs}
\begin{figure}[tbp]
    \centering
    \includegraphics[width=0.7\textwidth]{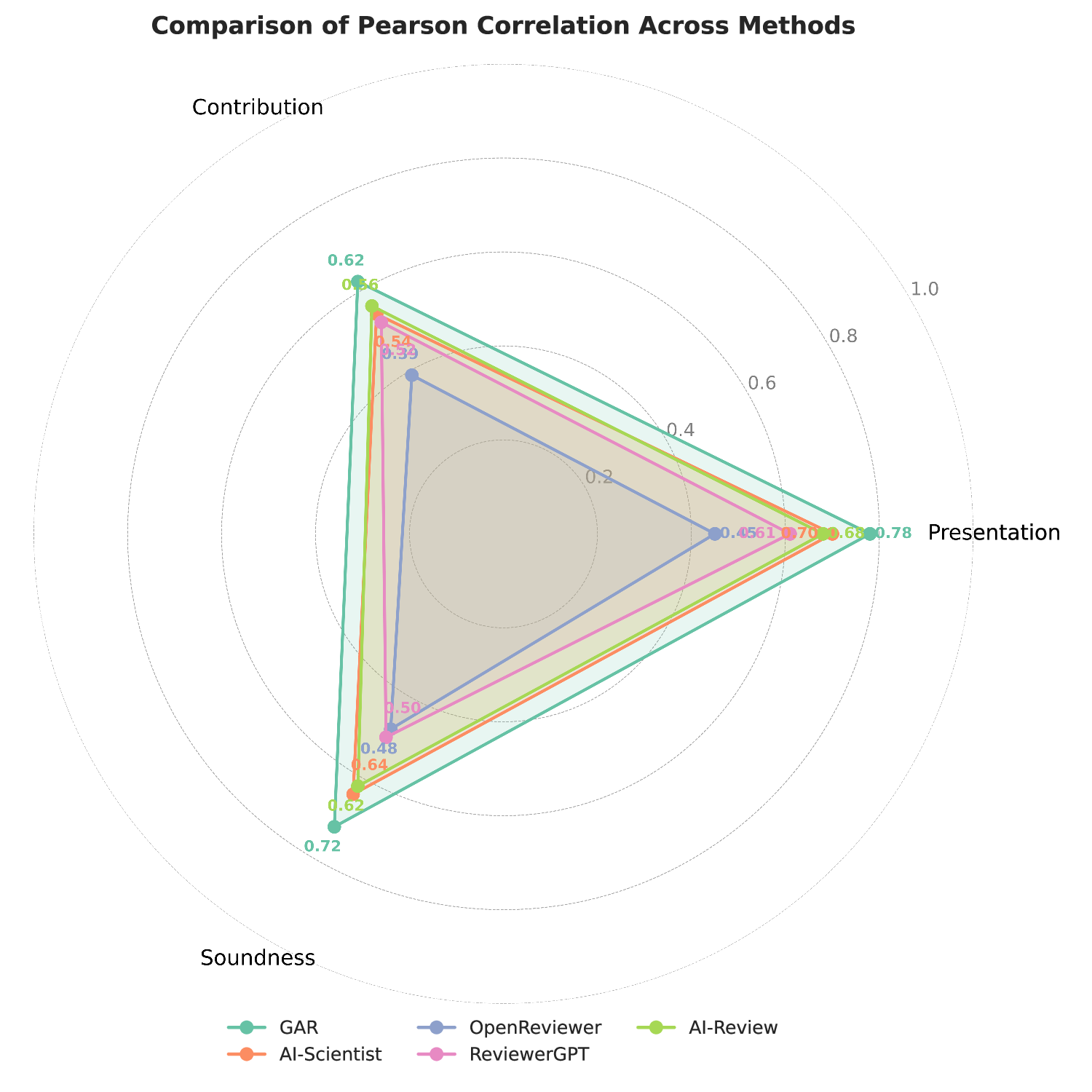}
    \caption{Alignment Between Human and LLM Reviewer Scores.}
    \label{fig:aligment}
\end{figure}
This experiment examines the alignment between the LLM-based reviewer and human reviewers on key review criteria: \textit{soundness}, \textit{presentation}, and \textit{contribution}. A subset of 1,000 papers from ICLR 2023 was selected, with both human and LLM-based reviewers providing scores for each criterion on a scale from 1 (Poor) to 4 (Excellent). Pearson correlation coefficients were calculated to measure the alignment of scores between human and LLM reviewers.

As illustrated in Figure \ref{fig:aligment}, GAR demonstrates high correlation coefficients across all criteria, suggesting that the LLM-based reviewer aligns closely with human reviewers. This strong alignment, especially on aspects of soundness and presentation, highlights the LLM's ability to approximate human assessment. While the correlation for the contribution attribute is comparatively lower, our approach still surpasses prior work. This may be attributed to the inherent challenge of assessing a paper's novelty based solely on limited contextual information, whereas human reviewers benefit from extensive field-specific expertise and years of experience.

\subsection{Human Likeliness}

LLM Evaluators \cite{chiang2023can} have shown that LLMs, such as GPT-4o, can serve as reliable evaluators, achieving performance comparable to human experts. In this experiment, we leverage GPT-4o to assess the feedback generated by LLM-based agents in the context of peer review. Specifically, we collect agent-generated reviews and prompted GPT-4o to evaluate whether these reviews were AI-generated or human-like. A 5-point Likert scale was used, where higher scores indicate a stronger resemblance to human reviewers’ style and consistency.

As shown in Table \ref{tab:llm_evaluator}, our method significantly outperforms AI-Scientist in generating reviews that align closely with human feedback. GAR scores are consistently higher, suggesting that the inclusion of graph-based memory and profile modules enhances the human-likeness of reviews. Additionally, allowing agents to simulate reviewer-specific characteristics, such as self-assessed confidence and depth of expertise, further contributed to review consistency and believability. Conversely, OpenReviewer and ReviewerGPT displayed tendencies towards inconsistency, such as generating shallow comments or narrowly focusing on methodological details without evaluating the validity of technical claims and results. This lack of critical depth raised suspicions of AI involvement.

\begin{table}[h]
\centering
\caption{Human-Likeness Scores by Reviewer Agent and Dataset}
\label{tab:llm_evaluator}
\begin{tabular}{lccc}
\toprule
 & \textbf{NeurIPS} & \textbf{ICLR 22} & \textbf{ICLR 23} \\
\midrule
AI-Scientist & \underline{3.34 $\pm$ 0.09}  & 3.39 $\pm$ 0.11 & \underline{3.38 $\pm$ 0.08}   \\
OpenReviewer & 2.45 $\pm$ 0.10 & 2.43 $\pm$ 0.09 & 2.43 $\pm$ 0.09   \\
ReviewerGPT & 3.26 $\pm$ 0.13 & 3.25 $\pm$ 0.14  & 3.29 $\pm$ 0.15   \\
AI-Review & 3.30 $\pm$ 0.09 & \underline{3.42 $\pm$ 0.11} & \underline{3.38 $\pm$ 0.08}   \\
\rowcolor{blue!10}
GAR & \textbf{3.89 $\pm$ 0.11}* & \textbf{4.02 $\pm$ 0.10}* & \textbf{3.99 $\pm$ 0.09}*  \\
\bottomrule
\end{tabular}
\end{table}

\subsection{Comparison of Feedback Between Synthetic Reviewers and Humans}

\begin{figure}[tbp]
    \centering
    \includegraphics[width=0.8\textwidth]{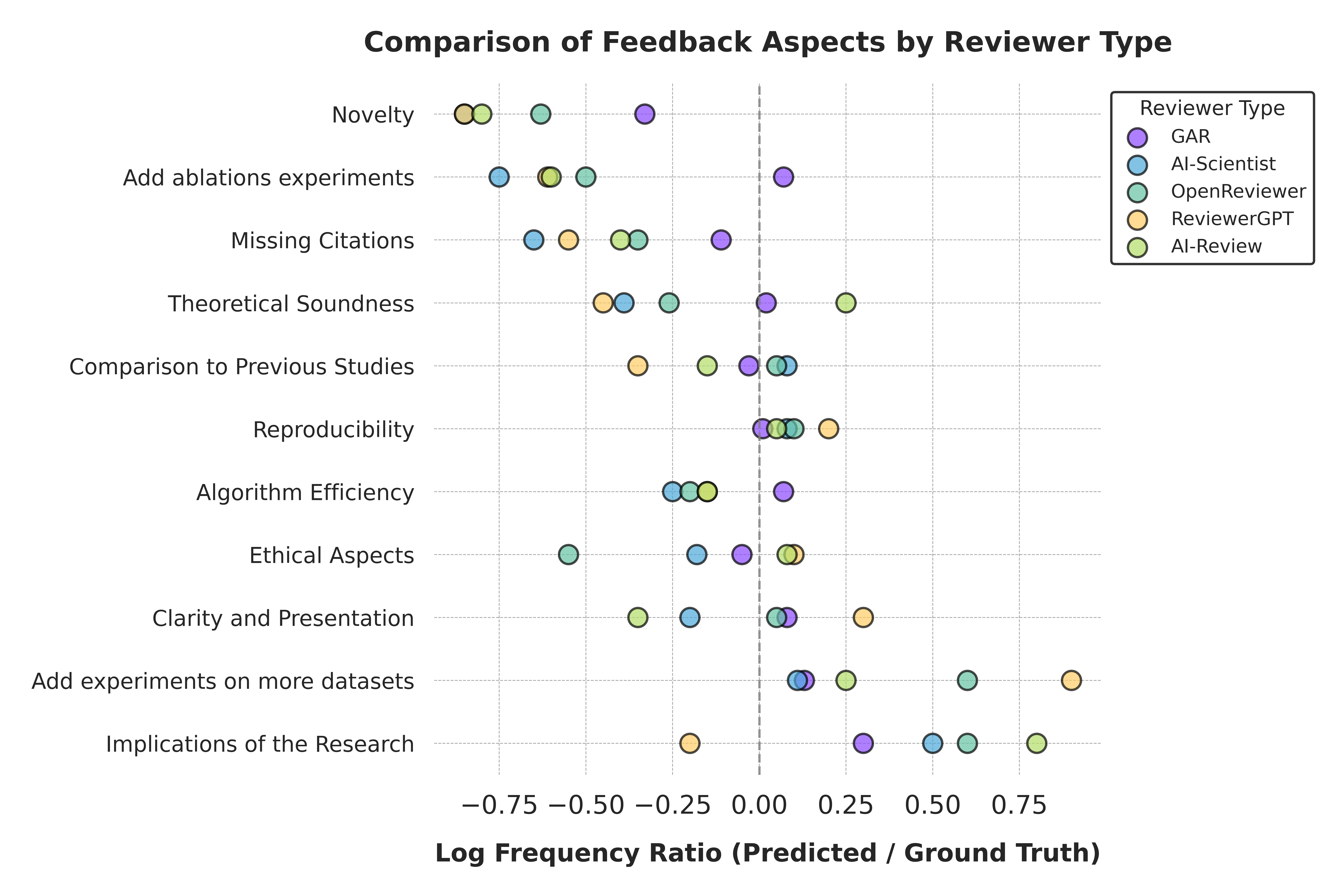}
    \caption{Relative frequency of feedback aspects by LLM, human reviewers, Method A, and Method B.}
    \label{fig:comparison}
\end{figure}

To understand whether some aspects of reviews are more/less likely to be discussed by agent and human reviewers, we analyze 11 aspects of comments. Human annotation was performed a randomly sampled subset of feedback, following established research in machine learning peer review \cite{birhane2022values, smith2022real}. 

Figure \ref{fig:comparison} displays the relative frequency of each feedback aspect for all six reviewer types: GAR, AI-Scientist, OpenReviewer, ReviewerGPT, and AI-Review. Notably, GAR emphasizes \textit{Implications of the Research} with a log frequency ratio of 0.3, indicating a slightly higher likelihood compared to human reviewers, and maintains moderate alignment for \textit{Add experiments on more datasets} (0.13). However, GAR underrepresents aspects such as \textit{Missing Citations} (-0.11) and \textit{Novelty} (-0.33), suggesting a more cautious focus on experimental and technical aspects over broad or less measurable categories. 

AI-Scientist demonstrates a strong focus on \textit{Implications of the Research} (0.51) but shows limited attention to \textit{Clarity and Presentation} (-0.22) and \textit{Theoretical Soundness} (-0.39). OpenReviewer, in contrast, highlights \textit{Add experiments on more datasets} (0.61) and \textit{Implications of the Research} (0.60), while underemphasizing \textit{Add ablations experiments} (-0.54) and \textit{Novelty} (-0.63). ReviewerGPT prioritizes \textit{Add experiments on more datasets} with the highest log frequency ratio (0.91) and provides moderate emphasis on \textit{Clarity and Presentation} (0.30), but significantly underrepresents \textit{Novelty} (-0.85). AI-Review shares similarities with OpenReviewer, placing strong emphasis on \textit{Implications of the Research} (0.82) and moderate focus on \textit{Add ablations experiments} (0.26), while underrepresenting \textit{Theoretical Soundness} (-0.38) and \textit{Novelty} (-0.79).

Taking advantage of its graph-based representation, GAR aligns more closely with human reviewers than prior work, particularly in technical domains such as experiments and results. While AI-Scientist and OpenReviewer frequently highlight experimental recommendations, GAR offers a more balanced assessment, closely aligning with human reviewers on \textit{Add ablations experiments} and other nuanced methodological aspects.

\subsection{Human Likeness Across Foundation Models}
\label{sec:human_like}

\begin{table}[h]
\centering
\caption{Human likeness with different types of foundation LLMs.}
\label{tab:llm_evaluator_likeliness}
\begin{tabular}{lccc}
\toprule
 & \textbf{NeurIPS} & \textbf{ICLR 22} & \textbf{ICLR 23} \\
\midrule
GPT-4o & \textbf{3.91 $\pm$ 0.10}  & \textbf{4.08 $\pm$ 0.10} & \textbf{4.11 $\pm$ 0.10}   \\
Mistral-7b Instruct & 3.59 $\pm$ 0.08 & 3.67 $\pm$ 0.10 & 3.68 $\pm$ 0.11   \\
Llama-3.1 (8b) & 3.33 $\pm$ 0.08  & 3.64 $\pm$ 0.11  & 3.64 $\pm$ 0.12   \\
Llama-3.1 (70b) & 3.66 $\pm$ 0.09  & 3.63 $\pm$ 0.10 & 3.73 $\pm$ 0.07   \\
\rowcolor{blue!10}
GPT-4o-mini & \underline{3.89 $\pm$ 0.11} & \textbf{4.02 $\pm$ 0.10} & \textbf{3.99 $\pm$ 0.09}  \\
\bottomrule
\end{tabular}
\end{table}

We assess the human-likeness of reviews generated by various foundation models, including GPT-4o, GPT-4o-mini, Mistral-7b Instruct, Llama-3.1 (8b), and Llama-3.1 (70b). The results, shown in Table \ref{tab:llm_evaluator_likeliness}, demonstrate that GPT-4o achieves the highest overall scores across all datasets, with a mean score of 4.11 $\pm$ 0.10 on ICLR 2023. This highlights GPT-4o’s ability to closely mimic human review styles. GPT-4o-mini, while slightly trailing GPT-4o on some datasets, performs competitively (e.g., 3.99 $\pm$ 0.09 on ICLR 2023) and offers a computationally efficient alternative, making it suitable for scenarios with limited resources.

Although Mistral-7b Instruct provides reasonable performance, achieving a mean score of 3.68 $\pm$ 0.11 on ICLR 2023, its scores indicate room for improvement compared to GPT-4-based models. Llama-3.1 (8b) performs similarly to Mistral-7b, with a score of 3.64 $\pm$ 0.12 on ICLR 2023. Interestingly, Llama-3.1 (70b), despite its larger size, does not outperform the smaller models consistently, suggesting that model scale does not necessarily correlate with human-likeness in review generation. Overall, GPT-4o and GPT-4o-mini emerge as the most human-like in their feedback across datasets, particularly on ICLR 2023.

\subsection{Acceptance Prediction Across Foundation Models}

\begin{table}[tbp]
\centering
\caption{Performance comparison of GAR and baselines on three datasets, each consisting of 1,000 papers. Results are presented for different foundation models.}
\resizebox{1.0\linewidth}{!}{
\begin{tabular}{lcccccc}
\toprule
\textbf{Methods} & \multicolumn{2}{c}{\textbf{NeurIPS}} & \multicolumn{2}{c}{\textbf{ICLR 22}} & \multicolumn{2}{c}{\textbf{ICLR 23}} \\
 & \textbf{Balanced Acc. $\uparrow$} & \textbf{F1 Score $\uparrow$} & \textbf{Balanced Acc. $\uparrow$} & \textbf{F1 Score $\uparrow$} & \textbf{Balanced Acc. $\uparrow$} & \textbf{F1 Score $\uparrow$} \\
\midrule
GPT-4o & \textbf{0.73 $\pm$ 0.03} & \textbf{0.71 $\pm$ 0.03} & \textbf{0.75 $\pm$ 0.03} & \textbf{0.72 $\pm$ 0.03} & \textbf{0.74 $\pm$ 0.03} & \textbf{0.73 $\pm$ 0.03} \\
GPT-4o$^{>}$ & \underline{0.70 $\pm$ 0.03} & \underline{0.68 $\pm$ 0.03} & \underline{0.73 $\pm$ 0.03} & \underline{0.70 $\pm$ 0.03} & \underline{0.72 $\pm$ 0.03} & \underline{0.71 $\pm$ 0.03} \\
\midrule
Mistral-7b Instruct & 0.62 $\pm$ 0.04 & 0.60 $\pm$ 0.04 & 0.64 $\pm$ 0.04 & 0.62 $\pm$ 0.05 & 0.65 $\pm$ 0.04 & 0.63 $\pm$ 0.05 \\
Mistral-7b Instruct$^{>}$ & 0.66 $\pm$ 0.05 & 0.64 $\pm$ 0.04 & 0.68 $\pm$ 0.04 & 0.66 $\pm$ 0.04 & 0.69 $\pm$ 0.05 & 0.67 $\pm$ 0.05 \\
\midrule
Llama-3.1 (8b) & 0.60 $\pm$ 0.06 & 0.58 $\pm$ 0.05 & 0.62 $\pm$ 0.05 & 0.60 $\pm$ 0.05 & 0.63 $\pm$ 0.04 & 0.61 $\pm$ 0.05 \\
Llama-3.1 (8b)$^{>}$ & 0.61 $\pm$ 0.04 & 0.59 $\pm$ 0.05 & 0.63 $\pm$ 0.05 & 0.61 $\pm$ 0.05 & 0.64 $\pm$ 0.04 & 0.62 $\pm$ 0.05 \\
\midrule
Llama-3.1 (70b) & 0.63 $\pm$ 0.03 & 0.61 $\pm$ 0.04 & 0.65 $\pm$ 0.07 & 0.63 $\pm$ 0.04 & 0.66 $\pm$ 0.03 & 0.64 $\pm$ 0.07 \\
Llama-3.1 (70b)$^{>}$ & 0.67 $\pm$ 0.04 & 0.65 $\pm$ 0.04 & 0.69 $\pm$ 0.04 & 0.67 $\pm$ 0.03 & 0.70 $\pm$ 0.05 & 0.68 $\pm$ 0.06 \\
\midrule
\rowcolor{blue!10}
GPT-4o-mini & 0.64$\pm$0.05 & 0.61$\pm$0.04 & 0.68$\pm$0.03 & 0.66$\pm$0.05 & 0.66$\pm$0.04 & 0.60$\pm$0.04 \\
\rowcolor{blue!10}
GPT-4o-mini$^{>}$ & 0.68$\pm$0.05 & 0.62$\pm$0.05 & 0.71$\pm$0.04 & 0.67$\pm$0.06 & 0.70$\pm$0.05 & 0.69$\pm$0.05 \\
\bottomrule
\end{tabular}
}
\label{fig:foundation_model}
\end{table}

We now seek to evaluate the performance of our methodology using various foundation models on the acceptance prediction task. Specifically, we compare the results obtained by employing GPT-4o-mini, GPT-4o, Mistral-7b Instruct, Llama-3.1 (8b) and Llama-3.1 (70b). The results, presented in Table \ref{fig:foundation_model}, demonstrate that the performance of GAR is generally robust across different foundation models. While GPT-4o exhibits significantly higher F1-score scores (t-test $p<0.05$), GPT-4o-mini achieves similar performance but with a lower inference time. Mistral-7b Instruct also performs reasonably well on the ICLR dataset. Among the smaller models, Mistral-7b Instruct$^{>}$ shows notable improvements, making it a competitive option for resource-constrained applications. However, Llama-3.1 models, particularly the 70b variant, demonstrate only modest gains despite their larger size, indicating diminishing returns for increased model complexity in this specific task, similarly to results obtained in Section \ref{sec:human_like}.

\section{Discussion}
In this work, we present GAR, one of the first framework for simulating the peer review process through the use of LLM-empowered agents. GAR agents autonomously analyze the manuscripts, evaluate their content, provide feedback, and predict acceptance outcomes. This end-to-end framework integrates stages of novelty assessment, multi-round review, and meta-review, aiming to replicate genuine reviewers in an cost-efficient and scalable manner. As a demonstration, the proposed method has been applied to major machine learning conferences, showcasing its potential to provide human-like feedback and determine which papers meet the quality standards for publication.

We also acknowledge that our method has certain limitations. One remaining challenge is identifying genuinely groundbreaking or paradigm-shifting ideas. GAR presents a novelty module that leverages external knowledge to detect innovative contributions at the paper-level. However, future work should focus on equipping synthetic reviewers with the ability to recognize novelty at a more nuanced level. This may include leveraging the knowledge graph structure of manuscripts to assess paper novelty at a community level, or using citation embeddings to capture shifts in research topics and trends \cite{shibayama2021measuring}. 

Despite efforts to reduce bias, AI models like GAR are not immune to inherent biases present in training data, which can impact the evaluation process and potentially disadvantage certain research fields or authors \cite{10.1162/coli_a_00524}. Establishing clear guidelines for ethical AI evaluation, incorporating fairness checks, regular audits, and opportunities for human oversight will be critical for maintaining trust in AI-driven review processes \cite{HAFFAR2019670}. Furthermore, we must question: \textit{Are we certain that these papers are not already part of the LLMs’ training corpus?} If such overlap exists, it could inadvertently introduce bias, as the model may demonstrate familiarity with the content, concepts, or style of certain papers, providing an unfair advantage or skewing evaluations. This issue is particularly pronounced for widely circulated preprints or seminal works that are likely to have influenced the training datasets of LLMs. Addressing this challenge presents a promising direction for future research. 

Another area of improvement lies in the memory module. Currently, it is initialized with a limited set of papers, which may restrict the diversity and depth of contextual references available to reviewer agents. Expanding this module with a larger dataset of reviews \cite{su2024two} could enhance the relevance and accuracy of retrieved feedback, allowing agents to identify more nuanced patterns and align their assessments more closely with human standards, but we leave it to future work to explore this direction further.

The credibility of peer review relies on the dedication, impartiality, and expertise of reviewers. Knowledge and subject-matter expertise are essential for accurately evaluating a paper's novelty, significance, and technical rigor. Fair intent helps to uphold trust by ensuring unbiased reviews, while a strong commitment from reviewers ensures a thorough evaluation process \cite{tennant2017multi}. However, given that peer reviewing is typically unpaid and time-intensive, this demanding role can sometimes lead to superficial assessments, potentially impacting the consistency and depth of reviews. This limitation is especially pertinent when designing synthetic review systems like GAR, which must capture these qualities to provide evaluations that reflect human-like consistency and insight. Addressing this challenge requires refining GAR's profile module to better emulate domain-specific expertise, fairness, and thoroughness in its assessments.

Our framework is designed to closely mirror real-world peer review practices, providing insights that are both relevant and credible. While GAR leverages LLMs to generate simulated reviews, ethical concerns arise when considering their use in actual review processes. Some conferences have reported an increase in AI-generated reviews \cite{yu2024is}, raising questions about the appropriateness of LLMs in authentic peer assessments. Although LLM-generated reviews can offer supplementary feedback, we advise against their use as replacements for human reviewers. Given the limitations of current LLMs, human oversight remains crucial to maintaining fairness, rigor, and integrity in peer review, ensuring quality evaluations aligned with academic standards.

We developed an automated paper review framework, which demonstrates that LLMs, while still evolving, can provide review quality close to human standards. GPT-4o consistently produced the best results, occasionally reaching scores above human experts. However, we do not rely solely on proprietary models; as LLMs advance, both open and closed models are likely to improve. Our approach, therefore, remains model-agnostic, balancing the high performance of closed models like GPT-4o with the flexibility, lower cost, and transparency of open models such as Llama-3. Although open models currently show slightly lower quality, they hold the potential for cost-effective and adaptable AI systems. Future efforts will explore a closed-loop, self-improving system using open models to maximize discovery potential.

\section{Conclusion}
This research marks a step towards improving scientific writing and research by offering cost-effective, in-depth, and on-demand reviews. We describe an architecture for generative reviewers that employs a graph-based representation of manuscripts, a memory module for storing past review experiences, and a novel technique to assign reviews with specific traits and preferences. We then demonstrate the potential of generative reviewers to achieve human-level feedback and accurately predict acceptance outcomes. Our vision is not to replace human reviewers but to enhance the review process by supporting them with synthetic reviewers capable of managing the increasing volume of submissions and providing early, constructive feedback. We also believe that this tools could be used to enhance the quality of research papers. This collaboration between AI-driven agents and human experts has the potential to accelerate scientific progress and empower researchers to focus on more ambitious challenges. Future work will address the nuances of open-ended evaluation, such as identifying novel ideas and minimizing biases, as we continue to refine GAR’s capabilities and explore its role in broader scientific ecosystems.

\bibliography{iclr2025_conference}
\bibliographystyle{iclr2025_conference}

\clearpage

\end{document}